\documentclass{article}
\usepackage{spconf,amsmath,graphicx}
\usepackage{amssymb}
\usepackage{amsthm}
\usepackage{booktabs}
\usepackage{algorithm}
\usepackage{algorithmic}
\usepackage{dsfont}
\usepackage{multirow}
\usepackage{bm}
\usepackage{cite}
\usepackage{pifont}

\title{Jointly Visual- and Semantic-Aware Graph Memory Networks for Temporal Sentence Localization in Videos}
\name{Daizong Liu$^1$  and Pan Zhou$^2\sthanks{Corresponding author.}$}
\address{$^1$Peking University \quad $^2$Huazhong University of Science and Technology}
\begin{document}
%
\maketitle
\begin{abstract}
Temporal sentence localization in videos (TSLV) aims to retrieve the most interested segment in an untrimmed video according to a given sentence query.
However, almost of existing TSLV approaches suffer from the same limitations:
(1) They only focus on either frame-level or object-level visual representation learning and corresponding correlation reasoning, but fail to integrate them both;
(2) They neglect to leverage the rich semantic contexts to further benefit the query reasoning.
To address these issues, in this paper, we propose a novel Hierarchical Visual- and Semantic-Aware Reasoning Network (HVSARN), which enables both visual- and semantic-aware query reasoning from object-level to frame-level.
Specifically, we present a new graph memory mechanism to perform visual-semantic query reasoning: For visual reasoning, we design a visual graph memory to leverage
visual information of video; For semantic reasoning, a semantic graph memory is also introduced to explicitly leverage semantic knowledge contained in the classes
and attributes of video objects, and perform correlation reasoning in the semantic space.
Experiments on three datasets demonstrate that our HVSARN achieves a new state-of-the-art performance.
\end{abstract}
\section{Introduction}
\label{sec:intro}

\begin{figure}[t!]
\centering
\includegraphics[width=0.35\textwidth]{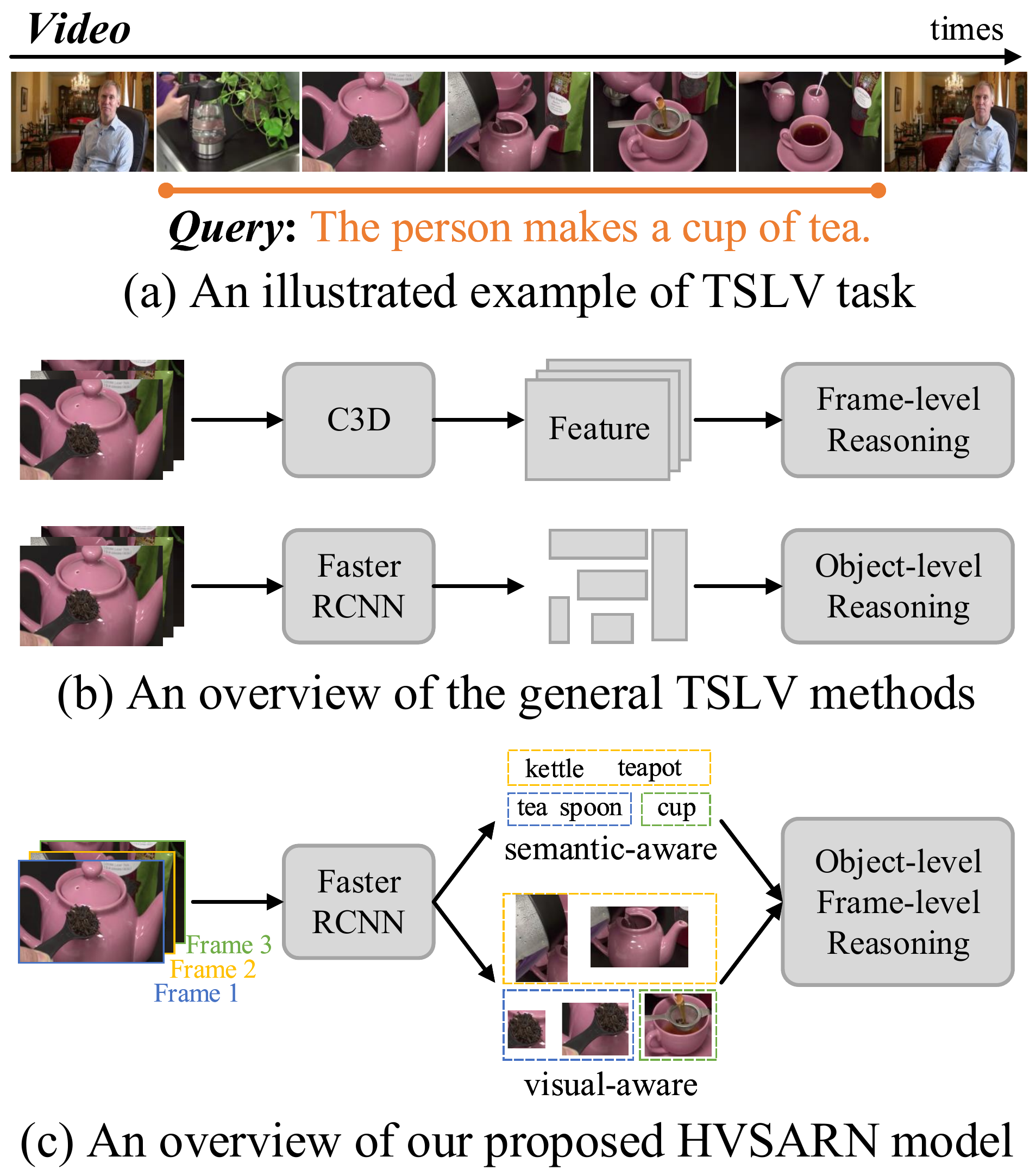}
\vspace{-16pt}
\caption{(a) An example of temporal sentence localization in videos (TSLV). (b) The general TSLV methods direcly extract frame-level or object-level visual features for query reasoning. (c) Our proposed method not only integrates both object- and frame-level visual features in a hierarchical way for learning more fine-grained contexts, but also leverages additional semantic information to assist the query reasoning.}
\label{fig:introduction}
\vspace{-16pt}
\end{figure}

Temporal sentence localization in videos (TSLV) is an important yet challenging task in natural language processing, which has drawn increasing attention over the last few years due to its vast potential applications in 
information retrieval and human-computer interaction. As shown in Figure \ref{fig:introduction} (a), it aims to ground the most relevant video segment according to a given sentence query. 
Most previous works \cite{liu2021context,anne2017localizing,chen2020learning,xu2021boundary,liu2022skimming,ji2021full,liu2022reducing,liu2023hypotheses,liu2022learning,fang2022hierarchical,fang2020v,fang2021unbalanced,fang2021animc,fang2022multi,fang2020double,fang2023you,xu2019mhp,liu2021f2net,liu2022imperceptible,liu2021spatiotemporal,liu2022rethinking,hu2022exploring} generally extract frame-level features of each video via a CNN network and then interact them with query for reasoning. Based on the cross-modal features, they either follow a proposal-based framework \cite{liu2022few,liu2020jointly} to select the best segment from multiple segment proposals, or follow a proposal-free framework \cite{fang2022multi,zhang2020span} to directly regress the temporal locations of the target segment.
However, their frame-level features may capture the redundant background information and fail to explore the fine-grained differences among video frames with high similarity.
Therefore, recently, some detection-based methods \cite{zeng2021multi,liu2023exploring} have been proposed to capture more fine-grained foreground object-level appearance features via a Faster RCNN model for more accurate query reasoning. The general pipeline of above methods are shown in Figure~\ref{fig:introduction} (b).

\begin{figure*}[t!]
\centering
\includegraphics[width=1.0\textwidth]{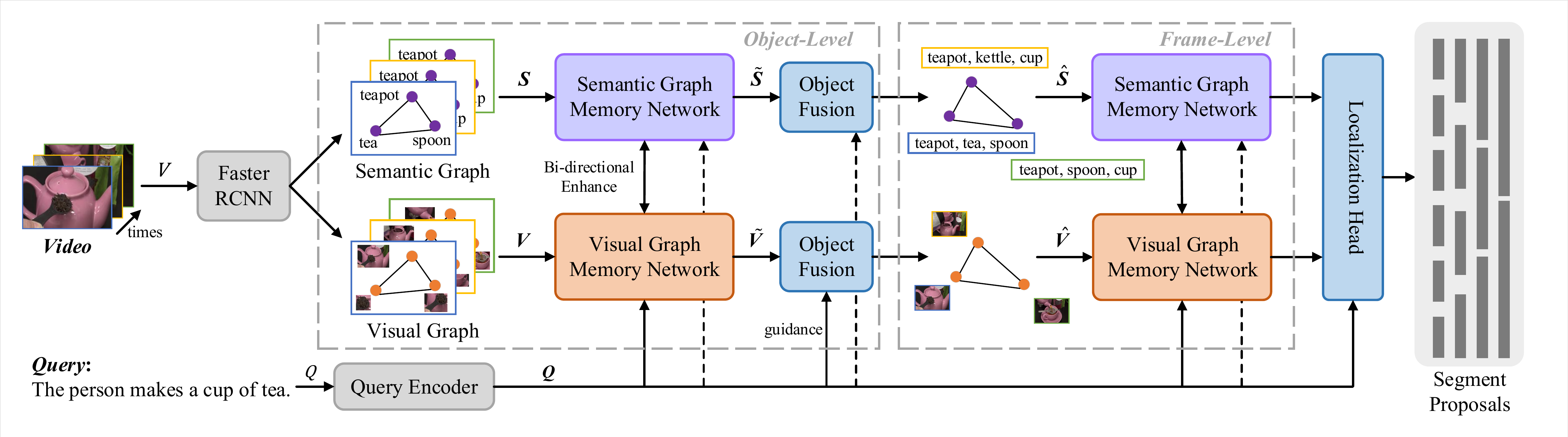}
\vspace{-24pt}
\caption{An overview of the HVSARN model. We first encode the input video into both visual and semantic spaces via a Faster-RCNN, and extract query information via a query encoder. Then, we construct visual and semantic graphs at object level for reasoning. After that, two object fusion modules are introduced to aggregate objects for each frame and build new graphs at frame level. At last, we utilize a localization head to predict the target segment.}
\label{fig:pipeline}
\vspace{-14pt}
\end{figure*}

Although these existing approaches have achieved promising results, they still suffer from two common limitations.
Firstly, current TSLV approaches only extract
frame-level or object-level features for reasoning, but fails to take the advantage of both of them.
Given a video-query pair, a typical and general localization
process is to first recognize query-relevant objects and their spatial relationships in each frame, and then reason the adjacent frames to model corresponding motions of their temporal relationship. 
However, none of the existing methods have developed their framework in a such hierarchical way (\textit{i.e.}, from object-level to frame-level).
Secondly, existing approaches only consider visual information for query reasoning, and neglect to utilize the additional semantic knowledge (\textit{e.g}, objects nouns) for assisting the reasoning process.
Detection-based methods only exploit object-level visual information using Faster RCNN, but neglect to leverage semantic knowledge (\textit{e.g.}, the attributes and classes of the detected objects) to capture more explicit and richer cues for benefiting the query reasoning.
For example, the simple action ``making tea" in Figure~\ref{fig:introduction} involves many commonsensible processes, such as boiling water, adding tea, and pouring it into a cup. It is crucial to capture these local semantic details (\textit{e.g.}, kettle, tea, teapot) among frames to comprehend the complete action ``making tea".

To this end, in this paper, we propose a novel Hierarchical Visual- and Semantic-Aware Reasoning Network (HVSARN), which jointly performs visual-semantic reasoning in a hierarchical structure.
Specifically, we achieve both visual and semantic reasoning by developing two types of graph memory networks \cite{liu2021hair}:
(1) We design a visual graph memory network to exploit visual object information of video, and gradually learn query-related visual representation for activity modelling; 
(2) We represent the classes and attributes of the detected objects as nodes and build edges to encode commonsense semantic relationships. A semantic graph memory network is built based on them to leverage semantic knowledge to facilitate query reasoning.
Both two graph memory mechanisms work cooperatively and interact with each other via learnable visual-to-semantic and semantic-to-visual enhancement.
To further enabling hierarchical visual-semantic reasoning, we build the HVSARN in a hierarchical structure from object-level to frame-level.

\section{The Proposed Method}
Given an untrimmed video $V$ and a sentence query $Q$, the task aims to determine the start and end timestamps $(s, e)$ of a specific video segment, which corresponds to the activity of the given sentence query. Formally, we represent the video as $V=\{v_t\}_{t=1}^T$ frame-by-frame where $T$ is the frame number, and denote the given sentence query as $Q=\{q_n\}_{n=1}^N$ word-by-word where $N$ is the word number. The details of our proposed method HVSARN are shown in Figure~\ref{fig:pipeline}.

\subsection{Video and Query Encoders}
\noindent \textbf{Video encoder.}
Given a video input $V$, we utilize a Faster R-CNN \cite{ren2015faster} pre-trained on the VGenome dataset \cite{krishna2017visual} to extract the visual features of $K$
objects from each frame.
Therefore, there are total $T \times K$ objects in a single video, and we can represent their visual features as $ \{\bm{o}_{t,k}, \bm{b}_{t,k}\}_{t=1,k=1}^{t=T,k=K}$, where $\bm{o}_{t,k} \in \mathbb{R}^{D}, \bm{b}_{t,k} \in \mathbb{R}^{4}$ denotes the feature and bounding-box position of the $k$-th object in $t$-th frame. 
Then, the visual object feature and the location feature are projected into same latent space with two learned linear layers, and are summed up as the initial object-level visual representation $\bm{V} = \{\bm{v}_{t,k}\}_{t=1,k=1}^{t=T,k=K}$.
In the meanwhile, we also extract the classes and attributes of the detected $K$ objects using the same Faster R-CNN. These semantic knowledge is embedded by a pre-trained fastText \cite{bojanowski2017enriching} model, and are then linearly projected into a $D$-dimensional space to produce the initial semantic representations $\bm{S} = \{\bm{s}_{t,k}\}_{t=1,k=1}^{t=T,k=K}$.

\noindent \textbf{Query encoder.}
Given the query input $Q$, we first utilize the Glove \cite{pennington2014glove} to embed each word into dense vector, and then employ multi-head self-attention and Bi-GRU \cite{chung2014empirical} to encode its sequential information. The final sentence-level feature $\bm{Q} \in \mathbb{R}^{D}$ can be obtained by concatenating the last hidden unit outputs in Bi-GRU.

\subsection{Object-Level Query Reasoning}
\noindent \textbf{Graph construction.}
After obtaining the visual embeddings $\bm{V}$, for each frame $t$, we take its objects $\{\bm{v}_{t,k}\}_{k=1}^K$ as nodes and build $t$-th visual graph by fully connecting them. We can also construct and initialize the semantic graphs based on the semantic embeddings $\bm{S} = \{\bm{s}_{t,k}\}_{t=1,k=1}^{t=T,k=K}$.

\noindent \textbf{Visual graph reasoning.}
We develop a graph memory controller in the visual graph to carry query information and interact it with the visual node representations by a series of read and write operations.
Specifically, we denote the initial state
of read controller as query feature $\bm{Q}$ and denote the initial representation of the $k$-th graph node in $t$-th frame as $\bm{v}_{t,k}$.
At each reasoning step $l \in \{1,2,...L\}$, the read controller attentively reads the content $\bm{r}^l$ from all nodes:
\begin{equation}
    \alpha_{t,k}^l = \bm{\text{w}}^{\top} tanh(\bm{W}_1^{\alpha}\bm{Q}^{l-1} + \bm{W}_2^{\alpha}\bm{v}_{t,k}^{l-1} + \bm{b}^{\alpha}),
\end{equation}
\vspace{-14pt}
\begin{equation}
    \bm{r}^l = \sum_{k=1}^K softmax(\alpha_{t,k}^l) \cdot \bm{v}_{t,k}^{l-1},
\end{equation}
where $\bm{W}_1^{\alpha},\bm{W}_2^{\alpha},\bm{b}^{\alpha}$ are learnable parameters, $\bm{\text{w}}$ is the row vector \cite{zhang2019cross}. Once acquiring node content $\bm{r}^l$, the read controller updates its state via a gate mechanism as:
\begin{equation}
    (\bm{Q}^{l-1})' = tanh(\bm{W}_1^r \bm{Q}^{l-1} + \bm{U}_1^r \bm{r}^l + \bm{b}_1^r),
\end{equation}
\vspace{-14pt}
\begin{equation}
    \bm{G}^l = sigmoid(\bm{W}_2^r \bm{Q}^{l-1} + \bm{U}_2^r \bm{r}^l + \bm{b}_2^r),
\end{equation}
\vspace{-14pt}
\begin{equation}
    \bm{Q}^l = \bm{G}^l \odot \bm{Q}^{l-1} + (1-\bm{G}^l) \odot (\bm{Q}^{l-1})',
\end{equation}
where $\bm{W}^r,\bm{U}^r,\bm{b}^r$ are the learnable parameters, $\odot$ denotes the element-wise multiplication. The update gate $\bm{G}^l$ controls how much previous state to be preserved.

After the read operation, we also need to update the node representations with new query information and the relations among nodes. At each step $l$, the write controller updates the $k$-th node in $t$-th frame by considering its previous representation $\bm{v}_{t,k}^{l-1}$, current query $\bm{Q}^l$ from the read controller, and the representations $\{\bm{v}_{t,i}^{l-1}\}_{i=1,i \neq k}^{i=K,i\neq k}$ from other nodes.
In detail, we first aggregate the information from neighbor nodes to capture the contexts:
\begin{equation}
    \bm{c}^l_{t,k} = \sum_{i=1,i\neq k}^{i=K,i\neq k}softmax(MLP([\bm{v}_{t,k}^{l-1},\bm{v}_{t,i}^{l-1}]))\bm{v}_{t,i}^{l-1},
\end{equation}
where $MLP$ is Multi-Layer Perceptron, $[,]$ denotes the concatenation. After obtaining the context
representation $\bm{c}^l_{t,k}$, the write controller updates the node representation via another gate function as:
\begin{equation}
    (\bm{v}^{l-1}_{t,k})' = tanh(\bm{W}_1^c \bm{v}^{l-1}_{t,k} + \bm{U}_1^c \bm{Q}^l + \bm{H}_1^c \bm{c}^l_{t,k} + \bm{b}_1^c),
\end{equation}
\vspace{-14pt}
\begin{equation}
    \bm{Z}^l = sigmoid(\bm{W}_2^c \bm{v}^{l-1}_{t,k} + \bm{U}_2^c \bm{Q}^l + \bm{H}_2^c \bm{c}^l_{t,k} + \bm{b}_2^c),
\end{equation}
\vspace{-14pt}
\begin{equation}
    \bm{v}^{l}_{t,k} = \bm{Z}^l \odot \bm{v}^{l-1}_{t,k} + (1-\bm{Z}^l) \odot (\bm{v}^{l-1}_{t,k})'.
\end{equation}


\noindent \textbf{Semantic graph reasoning.}
We also develop a graph memory controller in the semantic graph to leverage semantic knowledge to perform iterative reasoning over semantic representations.
Different from the visual graph, the semantic graph has additional updated representations of the visual graph $\{\bm{v}_{t,k}^L\}_{t=1,k=1}^{t=T,k=K}$ as input. Before the read controller updating, we first enhance the semantic representation of each node $\bm{s}_{t,k}$ using the visual evidence of its corresponding node from the visual graph as:
\begin{equation}
\label{eq:map1}
    \bm{f}_{t,k}^{vs} = \sum_{k=1}^{K}softmax(\bm{W}^f_1 [\bm{v}_{t,k}^{L},\bm{s}_{t,k}]) (\bm{W}^f_2 \bm{v}_{t,k}^{L}),
\end{equation}
where $\bm{W}_1^f$ is a trainable weight matrix to mapping the feature from the visual node to the semantic node, $\bm{W}_2^f$ is a projection weight matrix. The enhanced representation of each semantic node is denoted as: $\widetilde{\bm{s}}_{t,k} = [\bm{s}_{t,k},\bm{f}_{t,k}^{vs}]$.

Then, based on this initial semantic representations,  we perform iterative query reasoning. The corresponding read and write operations are the same as those in the visual graph. After $L$ reasoning steps, we obtain
the updated semantic representations $\widetilde{\bm{S}}^L=\{\widetilde{\bm{s}}_{t,k}^L\}_{t=1,k=1}^{t=T,k=K}$, which is then mapped back into visual space to further enrich the visual representation with global semantic knowledge via a semantic-to-visual enhancement:
\begin{equation}
\label{eq:map2}
    \bm{f}_{t,k}^{sv} = \sum_{k=1}^{K}softmax(\bm{W}^f_3 [\widetilde{\bm{s}}_{t,k}^L,\bm{v}_{t,k}^{L}]) (\bm{W}^f_4 \widetilde{\bm{s}}_{t,k}^L),
\end{equation}
where $\bm{W}_3^f,\bm{W}_4^f$ are learnable projection weights. The final enhanced representation of the $k$-th visual node is obtained using a residual connection: $\widetilde{\bm{v}}_{t,k} = [\bm{v}_{t,k}^{L},\bm{f}_{t,k}^{sv}]$.

\subsection{Frame-Level Query Reasoning}
\noindent \textbf{Object feature fusion.}
After obtaining the updated object-level features $\widetilde{\bm{V}}=\{\widetilde{\bm{v}}_{t,k}\}_{t=1,k=1}^{t=T,k=T}$ and $\widetilde{\bm{S}}=\{\widetilde{\bm{s}}_{t,k}\}_{t=1,k=1}^{t=T,k=T}$, we aim to integrate the objects within each frame to represent more fine-grained frame-level information under the guidance of query information, thus enabling subsequent frame-level query reasoning.
Specifically, for visual graph, we aggregate the nodes in each graph via a query-guided attention \cite{liu2022exploring}: $\widehat{\bm{v}}_t = Attn(\{\widetilde{\bm{v}}_{t,k}\}_{k=1}^K,\bm{Q})$, where $\widehat{\bm{v}}_t$ is the aggregated visual representation of the $t$-th frame and $\widehat{\bm{v}}_t \in \widehat{\bm{V}}$.
For semantic graph, we simply aggregate nodes using average pooling: $\widehat{\bm{s}}_t = avg(\{\widetilde{\bm{s}}_{t,k}\}_{k=1}^K)$, where $\widehat{\bm{s}}_t \in \widehat{\bm{S}}$.

\noindent \textbf{Visual-semantic graph reasoning.}
After obtaining the frame-level visual feature $\widehat{\bm{V}}$ and semantic feature $\widehat{\bm{S}}$, we construct two new fully-connected graphs based on them. Afterwards, both graph memory mechanisms perform iterative query reasoning over visual frame representations and semantic frame representation, respectively.

\begin{table*}[t!]
    \small
    \centering
    \setlength{\tabcolsep}{0.8mm}{
    \begin{tabular}{c|cccc|cccc|cccc}
    \hline \hline
    \multirow{3}*{Method} & \multicolumn{4}{c|}{ActivityNet Captions} & \multicolumn{4}{c|}{TACoS} & \multicolumn{4}{c}{Charades-STA} \\ \cline{2-5} \cline{6-9} \cline{10-13}
    ~ & R@1, & R@1, & R@5, & R@5, & R@1, & R@1, & R@5, & R@5, & R@1, & R@1, & R@5, & R@5, \\ 
    ~ & IoU=0.5 & IoU=0.7 & IoU=0.5 & IoU=0.7 & IoU=0.3 & IoU=0.5 & IoU=0.3 & IoU=0.5 & IoU=0.5 & IoU=0.7 & IoU=0.5 & IoU=0.7 \\ \hline
    CTRL & 29.01 & 10.34 & 59.17 & 37.54 & 18.32 & 13.30 & 36.69 & 25.42 & 23.63 & 8.89 & 58.92 & 29.57 \\
    ACRN & 31.67 & 11.25 & 60.34 & 38.57  & 19.52 & 14.62 & 34.97 & 24.88 & 20.26 & 7.64 & 71.99 & 27.79 \\
    QSPN & 33.26 & 13.43 & 62.39 & 40.78 & 20.15 & 15.23 & 36.72 & 25.30 & 35.60 & 15.80 & 79.40 & 45.40 \\
    CBP & 35.76 & 17.80 & 65.89 & 46.20 & 27.31 & 24.79 & 43.64 & 37.40 & 36.80 & 18.87 & 70.94 & 50.19 \\
    SCDM & 36.75 & 19.86 & 64.99 & 41.53 & 26.11 & 21.17 & 40.16 & 32.18 & 54.44 & 33.43 & 74.43 & 58.08 \\
    VSLNet & 43.22 & 26.16 & - & - & 29.61 & 24.27 & - & - & 54.19 & 35.22 & - & - \\
    CMIN & 43.40 & 23.88 & 67.95 & 50.73 & 24.64 & 18.05 & 38.46 & 27.02 & - & - & - & - \\
    2DTAN & 44.51 & 26.54 & 77.13 & 61.96 & 37.29 & 25.32 & 57.81 & 45.04 & 39.81 & 23.25 & 79.33 & 51.15 \\
    DRN & 45.45 & 24.36 & 77.97 & 50.30 & - & 23.17 & - & 33.36 & 53.09 & 31.75 & 89.06 & 60.05 \\
    CBLN & 48.12 & 27.60 & 79.32 & 63.41 & 38.98 & 27.65 & 59.96 & 46.24 & 61.13 & 38.22 & 90.33 & 61.69 \\
    MARN & - & - & - & - & 48.47 & 37.25 & 66.39 & 54.61 & 66.43 & 44.80 & 95.57 & 73.26 \\ \hline
    \textbf{HVSARN} & \textbf{55.76} & \textbf{34.29} & \textbf{87.62} & \textbf{70.31} & \textbf{51.85} & \textbf{41.04} & \textbf{69.41} & \textbf{57.93} & \textbf{69.62} & \textbf{47.99} & \textbf{97.35} & \textbf{76.40} \\ \hline \hline
    \end{tabular}}
    \vspace{-6pt}
    \caption{Performance compared with the state-of-the-arts on ActivityNet Caption, TACoS, and Charades-STA datasets.}
    \label{tab:compare}
    \vspace{-10pt}
\end{table*}

\subsection{Localization Head}
We integrate the updated frame-level visual and semantic features as the final multi-modal representation $\bm{M}=\{\bm{m}_{t}\}_{t=1}^T$, where $\bm{m}_t=[\widehat{\bm{v}}_t,\widehat{\bm{s}}_t]$.
With this feature, we further apply a bi-directional GRU network to absorb more contextual evidences in temporal domain. To predict the target video segment, we follow the same localization heads as \cite{zhang2019cross,liu2020jointly}.

\section{Experiments}
\subsection{Datasets and Evaluation Metrics}
We utilize three widely used benchmarks ActivityNet Caption \cite{krishna2017dense}, TACoS  \cite{regneri2013grounding}, and Charades-STA \cite{sigurdsson2016hollywood} for evaluation. we adopt ``R@n, IoU=m” as our evaluation metrics.

\subsection{Comparisons with the State-of-the-Arts}
Table \ref{tab:compare} summarizes the results on three challenging datasets. It shows that our proposed HVSARN outperforms all the baselines in all metrics.  Such significant improvement mainly attributes to the our additional usage of both semantic knowledge and hierarchical object-frame framework.

\begin{table}[t!]
    \small
    \centering
    \setlength{\tabcolsep}{1.0mm}{
    \begin{tabular}{l|cc|cc}
    \hline \hline
    \multirow{3}*{Setting} & \multicolumn{2}{c|}{TACoS} & \multicolumn{2}{c}{Charades-STA} \\ \cline{2-3} \cline{4-5}
    ~ & R@1, & R@5, & R@1, & R@5, \\ 
    ~ & IoU=0.5 & IoU=0.5 & IoU=0.7 & IoU=0.7 \\ \hline
    Object-level only & 38.73 & 54.61 & 45.02 & 73.36 \\
    Frame-level only & 34.48 & 50.19 & 41.17 & 68.54\\
    Two stream & 39.77 & 56.08 & 45.92 & 75.19 \\ \hline
    w/o visual & 35.78 & 52.25 & 42.10 & 69.14 \\
    w/o semantic & 38.82 & 55.17 & 45.91 & 73.59 \\
    w/o visual+semantic & 32.69 & 48.33 & 40.06 & 66.47 \\ \hline
    GCN & 39.24 & 55.57 & 46.18 & 73.86 \\
    GCN (fusion) & 40.63 & 56.90 & 47.15 & 75.52 \\
    Self-attention & 39.88 & 56.04 & 46.79 & 74.31 \\
    Memory network & 38.16 & 54.53 & 45.27 & 72.95 \\ \hline
    Full model & \textbf{41.04} & \textbf{57.93} &\textbf{47.99} & \textbf{76.40}
    \\ \hline \hline
    \end{tabular}}
    \vspace{-6pt}
    \caption{Ablation studies of our model.}
    \label{tab:ablation1}
    \vspace{-10pt}
\end{table}

\subsection{Ablation Study}
\noindent \textbf{Effect of hierarchical reasoning.}
We first conduct experiments to investigate the effect of the hierarchical reasoning framework. As shown in the first block of Table~\ref{tab:ablation1}, ablating any hierarchical level (\textit{i.e.}, object-level or frame-level) leads to severe performance degradation on all task types. 

\noindent \textbf{Effect of visual-semantic reasoning.}
We then analyze the impact of visual-semantic reasoning in the second block of Table~\ref{tab:ablation1}. It demonstrates that both visual and semantic contexts are crucial for the grounding performance.

\noindent \textbf{Effect of graph memory network.}
To investigate different variants of the graph memory network, we implement other reasoning modules in the third block of Table~\ref{tab:ablation1}. ``GCN" denotes the general graph convolutional network without memory mechanism, and ``GCN (fusion)" denotes the GCN module using additional fusion of multi-modal features as node representation.
These results demonstrate the superiority of our graph memory mechanism.

\begin{figure}[t!]
\centering
\includegraphics[width=0.48\textwidth]{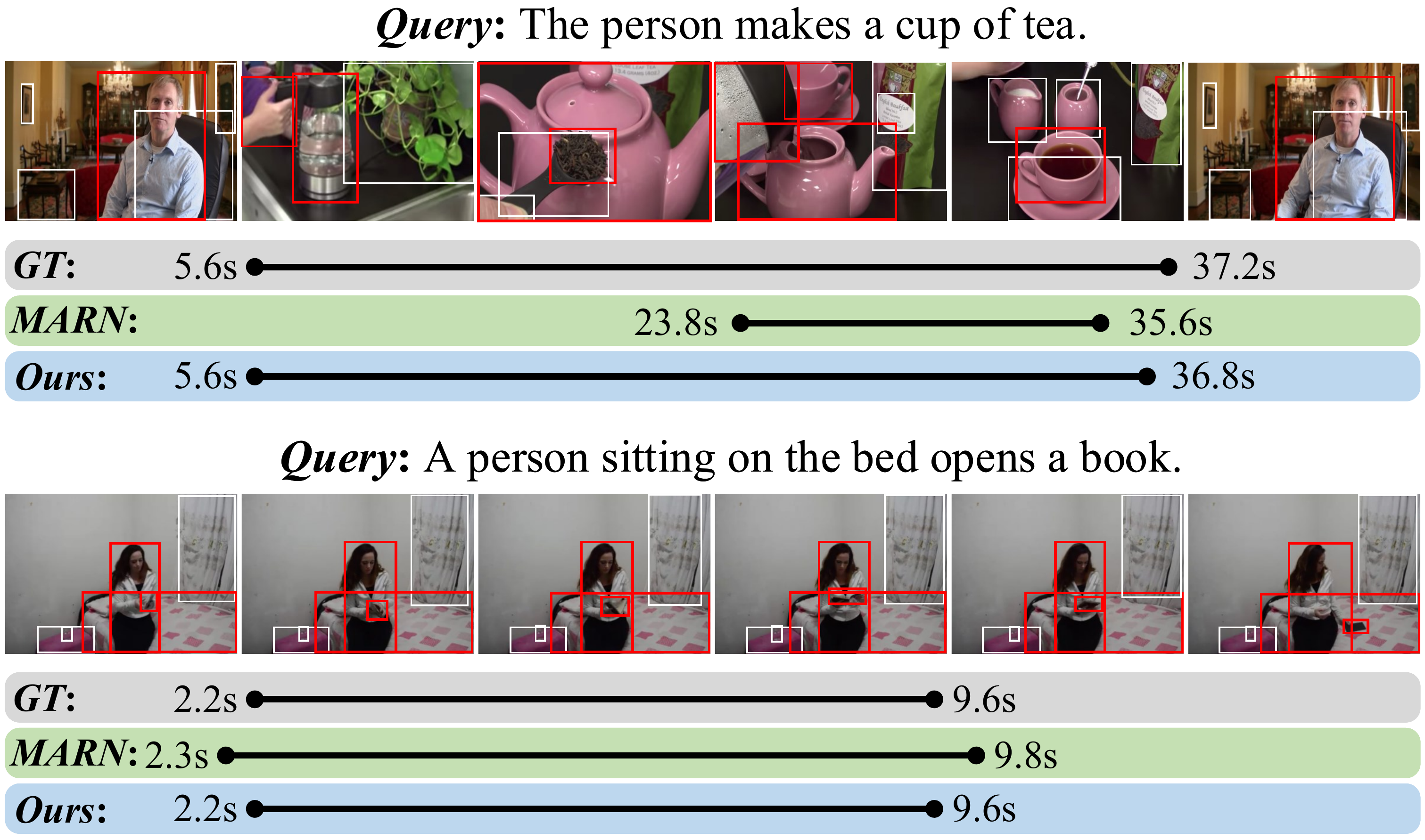}
\vspace{-20pt}
\caption{The qualitative results of the predicted segments.}
\label{fig:result}
\vspace{-16pt}
\end{figure}

\subsection{Visualization Results}
We provide two qualitative examples of our model and previous best detection-based MARN model in Figure~\ref{fig:result}. It shows that our method achieves better segment localization than MARN since we utilize additional semantic reasoning branch to fully comprehend the query.

\section{Conclusion}
In this paper, we propose a novel Hierarchical Visual- and Semantic-Aware Reasoning Network (HVSARN) for the TSLV task, 
which gradually focuses on spatial object-level reasoning to temporal frame-level reasoning in a hierarchical way. Extensive experiments conducted on three challenging datasets demonstrate the effectiveness of the proposed method.

\bibliographystyle{IEEEbib}
\bibliography{strings,refs}

\end{document}